\documentclass{article}

\usepackage{times}
\usepackage{graphicx} 
\usepackage{subfigure}

\usepackage{natbib}

\usepackage{algorithm}
\usepackage{algorithmic}

\usepackage{verbatim}

\usepackage{hyperref}

\usepackage[accepted]{icml2017}

\usepackage{hyperref}       
\usepackage{url}            
\usepackage{booktabs}       
\usepackage{amsfonts}       
\usepackage{bbm}
\usepackage{nicefrac}       
\usepackage{microtype}      
\usepackage{tikz}
\usepackage{amsmath}
\usepackage{amsfonts}
\usepackage{amssymb}
\usepackage{graphicx}
\usepackage{booktabs}
\usetikzlibrary{bayesnet} 
\usetikzlibrary{arrows}

\icmltitlerunning{Decoupling Dynamics and Reward}

\begin{document}

\twocolumn[
\icmltitle{Decoupling Dynamics and Reward for Transfer Learning}



\icmlsetsymbol{equal}{*}

\begin{icmlauthorlist}
\icmlauthor{Amy Zhang}{equal,to,fb}
\icmlauthor{Harsh Satija}{equal,to,fb}
\icmlauthor{Joelle Pineau}{to,fb}
\end{icmlauthorlist}

\icmlaffiliation{to}{McGill University}
\icmlaffiliation{fb}{Facebook AI Research}

\icmlcorrespondingauthor{Amy Zhang}{amy.x.zhang@mail.mcgill.ca}
\icmlcorrespondingauthor{Harsh Satija}{harsh.satija@mail.mcgill.ca}

\icmlkeywords{reinforcement learning, representation learning, transfer learning}

\vskip 0.3in
]



\printAffiliationsAndNotice{\icmlEqualContribution} 

\begin{abstract}
Current reinforcement learning (RL) methods can successfully learn single tasks, but often generalize poorly to modest perturbations in task domain or training procedure.  In this work we present a decoupled learning strategy for RL that creates a shared representation space where knowledge can be robustly transferred.  We separate learning the task representation, the forward dynamics, the inverse dynamics and the reward function of the domain, and show that this decoupling improves performance within task, transfers well to changes in dynamics and reward, and can be effectively used for online planning.  Empirical results show good performance in both continuous and discrete RL domains.
\end{abstract}

\section{Introduction}
\label{intro}
Reinforcement Learning (RL) provides a sound decision-theoretic framework to optimize the behavior of learning agents in an interactive setting. However, application of RL to real-world tasks is limited by several factors.  One challenge is the massive amounts of data required to learn an optimal behavior; this can be alleviated by using a high-fidelity simulator or game engine~\cite{openai16,elf17}, but there are many real-world domains where this is not available~\cite{kober13,shortreed11}.  Furthermore, RL policies trained within a simulator tend to overfit to the task, and generalize poorly even to modest perturbations in environment or task domain~\cite{henderson18}.  

The goal of our work is to design an RL model that can be efficiently trained on new tasks, and produce solutions that generalize well beyond the training environment. To do this, we adopt the framework of model-based RL~\cite{sutton90}.  We take particular inspiration from the work on Successor Features \cite{dayan1993improving}, which decouples the value function representation into dynamics and rewards, and learns them separately. In our work, we take this further and explicitly decouple learning the state representation, the reward function, the forward dynamics, and the inverse dynamics of the environment. We posit that we can learn a representation space $\mathcal{Z}$ via this decoupling that makes downstream learning easier. There are several reasons to pursue a decoupled approach: (1) the modules can be learned separately enabling efficient reuse of common knowledge across tasks to quickly adapt to new tasks; (2) the modules can be optimized jointly leading to a representation space that is adapted to the policy and value function, rather than the only the observation space; 
(3) the dynamics model enables forward search and planning, in the usual model-based RL way.  Our approach explicitly incorporates learning of inverse dynamics, and we show that this plays an important role in stabilizing learning. Empirical results confirm that learning in new domains can leverage this decomposition to achieve faster learning in a variety of domains, including continuous control MuJoCo tasks and discrete maze planning tasks.

\section{Technical Background}
Consider an RL agent deployed in a dynamic stationary environment.  The environment is modeled as a Markov Decision Process (MDP), which is defined by a set of states $\mathcal{S}$, a set of actions $\mathcal{A}$, dynamics $p(\cdot|s, a)$, and rewards $r(s, a)$.   The behavior of the RL agent is defined by a policy $\pi: \mathcal{S} \rightarrow \mathcal{A}$, specifying an action to apply in each state.  The goal is to learn an  optimal policy, denoted $\pi^*$, that maximizes the expected cumulative reward over trajectories.  The value function $V^\pi(s)$ and state-action value function $Q^\pi(s,a)$ are defined as usual in the RL literature~\cite{sutton1998reinforcement}.


Because our work is concerned with the robustness and generalizability of reinforcement learning agents, we also consider a distribution $\mathcal{D}$ over a family of tasks $\mathcal{T}$.  We define $\mathcal{T}$ to be the space of tasks that share $\mathcal{S}$ and $\mathcal{A}$, but dynamics $p(\cdot|s, a)$, and rewards $r(s, a)$ can vary.  We sample from $\mathcal{T}$ at training time. 
When the agent is in a particular task $\mathcal{T}_k$, it collects a set of trajectories, $D^{\mathcal{T}_k} = \{D^k_1, D^k_2, ..., D^k_n\}$, where $D^k_i=\{s_0, a_0, s_1, a_1, \dots, s_t, a_t, \dots, s_{T-1}, a_{T-1}, s_{T}\}$.

We consider specifically the case of model-based RL, where the dynamics and reward are estimated directly, and the optimal policy is found by applying dynamic programming on those quantities~\cite{sutton90,dyna}.  In the tabular (discrete state/action) case, the transition dynamics are estimated from state visitation counts and the reward function is estimated from expectation over training trajectories.  In more complex domains, the transition and reward functions can be estimated from richer regression models (see Sec.~\ref{related_work}).

\section{Decoupling model-based RL}

\begin{figure}[h]
  \centering
\includegraphics[width=0.75 \columnwidth]{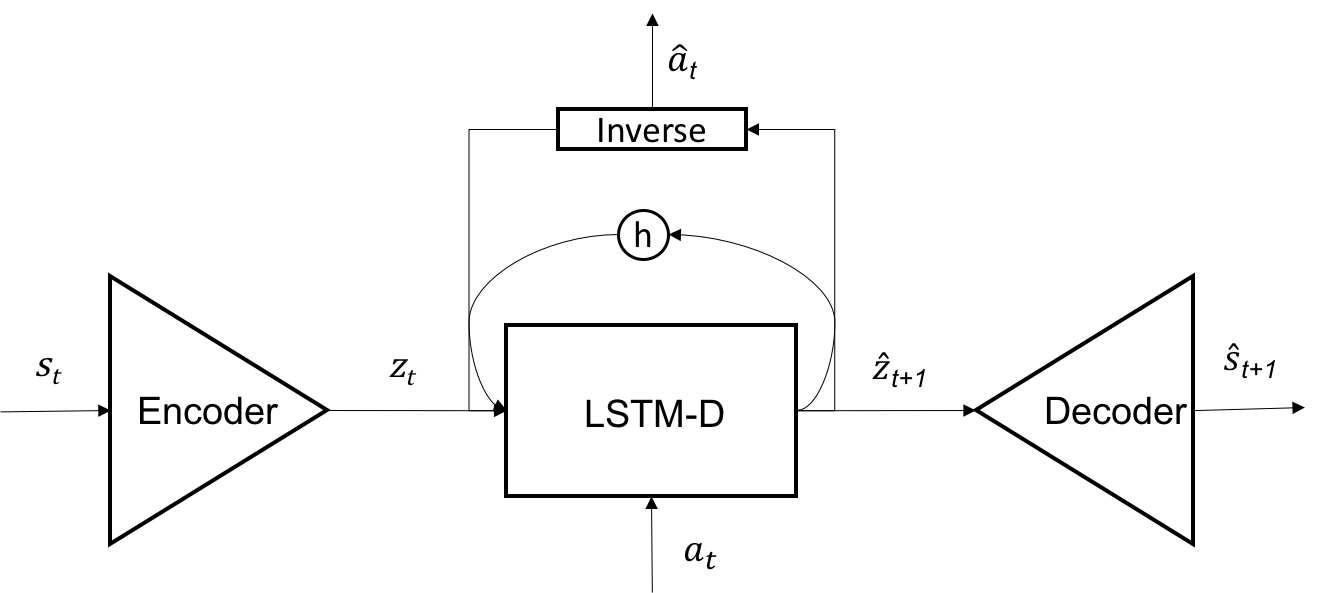}
  \caption{Dynamics Module}
\label{fig:dynamics}
\end{figure}
\begin{figure}[h]
  \centering
\includegraphics[width=0.75 \columnwidth]{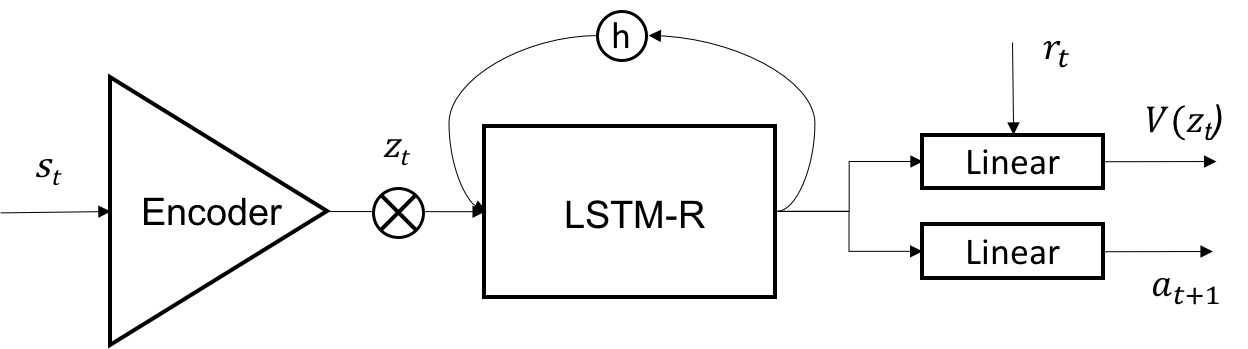}
  \caption{Rewards Module. $\otimes$ denotes the stop gradient operator, which doesn't allow the gradients to propagate back.}
\label{fig:rewards}
\end{figure}

Our objective is to provide a modular framework for model-based RL, leveraging a decomposition of the learning problem to provide reusable components that can be bootstrapped to enable fast re-training following changes in dynamics and rewards. The learning is decomposed into two complementary objectives, one for learning the state dynamics model and the other for learning the reward function.  Figures~\ref{fig:dynamics} \&~\ref{fig:rewards} give an overview of the proposed architecture. 
We define a learned representation space $\mathcal{Z}$ that we map to and from $\mathcal{S}$ with an encoder and decoder. It is through $\mathcal{Z}$ that our modules interface.

\subsection{A Modular Dynamics Model}

The goal of this first module is to learn the dynamics of the environment $p(\cdot|s, a)$.  
The main components of the dynamics module are an encoder $f_{enc}(s;\theta_{enc})$ and the forward model $f_{for}(z, a;\theta_{for})$. 
We add an additional two components: a decoder $f_{dec}(z;\theta_{dec})$ and inverse model $f_{inv}(z, z'; \theta_{inv})$.  These act as regularizers by enabling additional complementary training signals to be considered during learning. Since we are running our forward model in $\mathcal{Z}$ space, the decoder is necessary to evaluate our forward model. We also posit the inverse model is necessary as a constraint that causality is maintained in the representation space. Ablation experiments supporting these claims are in Section \ref{sec:ablations}.

The encoder and decoder pair allows us to learn a mapping between state space $\mathcal{S}$ and representation space $\mathcal{Z}$:
\begin{align}
  z_t &= f_{enc}(s_t; \theta_{enc}), \label{eq:encode}\\
  \hat{s}_t &= f_{dec}(z_t; \theta_{dec}). \label{eq:decode}
\end{align}

The forward model predicts transition probability $p(\cdot|s, a)$.  The inverse model observes the current and next state $z, z'\in\mathcal{Z}$, and aims to predict which action was taken to go from one to the other.  This can be an ill-posed problem, since more than one action can explain an observed transition.  Nonetheless we treat it as a supervised learning problem, since it is one component of a more complex optimization, and other terms restrict the solution space.

Traditionally, forward and inverse models map between state and action spaces $\mathcal{S}$, $\mathcal{A}$.
In our model, we learn the forward and inverse models in the learned representation space $\mathcal{Z}$.  Therefore, our models are of the form:
\begin{align*}
  \hat{z}_{t+1} &= f_{for}(z_t, a_t; \theta_{for}) \\
  \hat{a}_t &= f_{inv}(z_t, z_{t+1}; \theta_{inv})
\end{align*}
By abstracting away the model of dynamics to a representation space $\mathcal{Z}$, we have the freedom to encode more or less information than what exists in the given space $\mathcal{S}$.  We show that this abstraction allows for easier learning and improved results across a variety of environments.  

The forward model $f_{for}$ is learned using a recurrent architecture because we want the latent representation to incorporate temporal dependencies.  In doing so, we are relaxing the Markov Assumption on the model in the state space. This is useful because the latent state does not encode the exact observed state, thus the recurrent model can keep track of necessary past information to prevent aliasing. This has also been empirically shown to produce better results with A3C + LSTM \cite{mnih2016asynchronous}.  The recurrent next-state prediction takes the form:
\begin{align*}
    \hat{z}_{t+1}, h_{t} &= f_{for}(z_{t}, a_{t}, h_{t-1} ; \theta_{for}) \\
    \hat{s}_{t+1} &= f_{dec}({\hat{z}_{t+1}; \theta_{dec}})
\end{align*}
We instantiate $f_{for}$ with an LSTM \cite{Hochreiter:1997:LSM:1246443.1246450}. Here $h_t$ denotes the hidden state of the recurrent model.  Since we want the next latent state to generate the next observed state, we can use the reconstruction loss for the next state.

The total decoder loss, $\mathcal{L}_{dec}$, includes the reconstruction loss between $s_t$ and $\hat{s}_{t}$ (denoted $\mathcal{L}_{recon}$) as well as the next state prediction loss between $\hat{s}_{t+1}$ (predicted next state output of the forward model) and $s_{t+1}$ (denoted $\mathcal{L}_{state}(\theta_{for}, \theta_{dec})$). This second term folds in the effect of the forward model. 
\begin{align*}
   \mathcal{L}_{t, recon}(\theta_{enc}, \theta_{dec}) &= (\hat{s}_t - s_t)^2 \\
   \mathcal{L}_{t, state}(\theta_{for}, \theta_{dec}) &= (\hat{s}_{t+1} - s_{t+1})^2 \\
    \mathcal{L}_{t, dec}(\theta_{enc}, \theta_{dec}, \theta_{for}) &= \mathcal{L}_{recon} + \mathcal{L}_{state}.
\end{align*}

The forward model loss is similarly defined as:
\begin{align*}
  \mathcal{L}_{t, for}(\theta_{for}, \theta_{enc}) &= (\hat{z}_{t+1} -  z_{t+1})^2.
\end{align*}

The inverse model formulation and loss are defined as:
\begin{align*}
  \hat{a}_t \sim p(\hat{a}) &= f_{inv}(z_t, z_{t+1} ; \theta_{inv}), \\
  \mathcal{L}_{t, inv}(\theta_{inv}) &= -\sum_a p(a_t)\log(p(\hat{a}_t)),  \tag{discrete case} \\
  \mathcal{L}_{t, inv}(\theta_{inv}) &= (\hat{a}_t - a_t)^2.  \tag{continuous case}
\end{align*}
We use a cross-entropy loss for domains with discrete action spaces, and mean square error in continuous action cases.

Let $\theta_{dynamics} = \{ \theta_{inv}, \theta_{enc}, \theta_{dec}, \theta_{for}\}$, then the final loss for the trajectory can be written as :
\begin{align}
\label{eq:dynamics_loss}
  \mathcal{L}_{dynamics}(\theta_{dynamics}) &= \sum_{t=0}^{T} ( \lambda_{dec} \mathcal{L}_{t, dec} \\
    \qquad + \lambda_{for} \mathcal{L}_{t, for} & + \lambda_{inv}\mathcal{L}_{t, inv})
\end{align}
where $\lambda_{dec}$, $\lambda_{for}$, $\lambda_{inv}$ are (constant) hyper-parameters.

Note that this module learns a dynamics model purely with respect to trajectories; it ignores tasks and rewards.

\subsection{A Modular Reward Model}

Assuming the dynamics model outline above learns a latent representation that captures the dynamics, the goal of the Reward Module is to learn the value function and policy over this representation space (rather than in the raw state space). The reward module is the primary decision-making module -- it selects the next action and predicts the expected value. We use an Actor-Critic method~\cite{sutton2000policy} to learn the policy and value function simultaneously:
\begin{align*}
  \pi(a_t|z_t ; \theta_{actor}) &= f_{actor}(z_t; \theta_{actor}) \\
  V(z_t; \theta_{critic}) &= f_{critic}(z_t ; \theta_{critic})
\end{align*}
 using TD learning with multi-step bootstraps~\cite{sutton1988learning}. Let $R= r_t + \gamma V(z_{t+1}; \theta_{critic})$ be the estimated expected return, then the losses for the actor and critic are:
\begin{align}
  \mathcal{L}_{t, actor}(\theta_{actor}) &=  -\log \pi(a_t|z_t; \theta_{actor}) (R - V(z_{t}; \theta_{critic})) \\
  \mathcal{L}_{t, critic}(\theta_{critic}) &=  (R - V(z_{t}; \theta_{critic}))^2
\end{align}
This can also be extended to multi-step actions as in A3C~\cite{mnih2016asynchronous}. 

We can also have a recurrent formulation of actor-critic as:
\begin{align*}
  V(z_{t}), \pi(a_t|z_t), h_{t} &= f_{reward}(z_{t}, h_{t-1})
\end{align*}
Where $f_{reward}$ is the combination of $f_{actor}, f_{critic}$. Note that we do not necessarily need an LSTM for this module, but we believe that including history $H=\{s_1, a_1, r_1, s_2, a_2, r_2, ..., s_{t-1}, a_{t-1}, r_{t-1}\}$ often stabilizes training and incorporates extra information to the policy that is useful, even if technically unnecessary, as empirically shown in \citet{mnih2016asynchronous}.

The reward module is equivalent to classic actor-critic, except it learns the value function and policy in the representation space $\mathcal{Z}$ rather than in the state space $\mathcal{S}$.  We introduce another hyper-parameter $\lambda_{critic}$ to calibrate the effect of the critic loss relative to the actor loss.   Now our total loss for the reward module is 
\begin{align}
\label{eq:reward_loss}
\mathcal{L}_{reward}(\theta_{reward}) =  \sum_{t=0}^{T} (\lambda_{critic}\mathcal{L}_{t, critic} + \mathcal{L}_{t, actor})
\end{align}

\section{Inference in the Decoupled Model}
Our model accommodates both \textit{offline} and \textit{online} model learning, as well as both \textit{off-policy} and \textit{on-policy}\footnote{See~\citet{dyna} for standard definitions of these terms.}.  We also discuss how the dynamics model can be used for online planning.

\subsection{Dynamics Learning}
\label{sec:dynamics-learn}
\textit{Offline, Off-policy}. The dynamics module can be trained in a supervised manner and off-policy, since its goal is only to explore and learn the dynamics of the environment in a passive manner.  For the offline off-policy training case, we generate samples with an exploratory policy (usually uniform random action selection) and train the dynamics module using the sampled trajectories.  This is the most common mode of training the dynamics model, especially when task robustness and transfer are desired.  All modules in the dynamics model (encoder, decoder, forward, inverse) are jointly trained with the same set of batch samples, as per Algorithm \ref{alg:dynamics}. 
The main advantage of this approach is that we can bootstrap data collected from previous tasks, and having a batch of data from an exploratory policy generally leads to more stable learning, compared to on-policy training. Assuming that the policy used to collect the data is sufficiently exploratory, we are able to learn a representation space that  captures useful information for a family of tasks.  Clearly there is a trade-off here:  more exploration provides more robust information, but is less efficient than a narrowly targeted policy.

\textit{Offline, On-policy}.  Rather than using an exploratory policy, the dynamics model can be trained using the target policy.  This setup is less common, since having a good dynamics model is usually a precursor to a good policy in model-based RL.

\textit{Online, On-policy}.  The dynamics module can also be trained with samples drawn from the rewards module.  In this case the training happens online, through repeated interactions with the environment, and on-policy, through updates to the policy estimated in the reward module. This case is further detailed below in Sec.~\ref{sec:reward-learn} since both modules are trained simultaneously.

\textit{Online, Off-policy}.  Finally, we can train the dynamics model online, but using a policy different than the one learned by the reward module. For example, we can inject exploratory noise into the policy of the actor.  In this case we can improve training stability, at some extra cost in data acquisition.


\begin{figure}[h]
    \begin{algorithm}[H]
    \caption{Dynamics Training Algorithm}
    \label{alg:dynamics}
    \begin{algorithmic}
    \STATE Initialize module parameters $\theta_{dynamics}$
    \STATE Initialize hidden state $h_d$ for LSTM-D
    \STATE Set dynamics hyper-parameters $\lambda_{inv}, \lambda_{dec}, \lambda_{for}$
    \FOR{$e\in\{1,...,E\}$}
      \FOR{$(s_i, a_i, s_i') \in \{(s_0, a_0, s_0'),...,(s_N, a_N, s_N')\}$}
        \STATE Encode $s_i, s_i'$ to $z_i, z_i'$ (Eq. \ref{eq:encode})
        \STATE $\hat{z}_i'\leftarrow f_{for}(z_i, a)$ 
        \STATE $\hat{a}_i\leftarrow f_{inv}(z_i, z_i')$
        \STATE Decode $z_i, \hat{z}_i'$ to $\hat{s}_i, \hat{s}_i'$ (Eq. \ref{eq:decode})
        \STATE Compute $\mathcal{L}_{dynamics}$ (Eq. \ref{eq:dynamics_loss})
      \STATE Update $\theta_{dynamics}$
   \ENDFOR
    \ENDFOR
    \end{algorithmic}
    \end{algorithm}
\footnotesize{Alg.1 notes: $h_d$ is the hidden state of the LSTM in the dynamics module (LSTM-D). $E$ is the number of epochs, $N$ is the number of training samples.  We show the rollout 1 and batch size 1 case, but can be extended to longer rollouts where we see trajectories of length $r$ and compute and update in batches of size $b$ for speed.}
\end{figure}

\begin{figure}[h]
\begin{algorithm}[H]
\label{alg:reward}
\caption{Reward Training Algorithm}
\begin{algorithmic}
\STATE Freeze $\theta_{enc}$
\STATE Initialize module parameters $\theta_{reward}$
\STATE Initialize hidden state $h_r$ for LSTM-R
\STATE Set reward hyper-parameter $\lambda_{critic}$ 
\STATE $t\leftarrow 1$, $T\leftarrow 0$
\REPEAT
    \STATE Clear gradients
    \IF{episode done}
       \STATE Clear hidden states $h_r, h_d$
       \STATE Reset environment
    \ENDIF
    \STATE $t_{start}=t$
    \STATE Get state $s_t$
    \REPEAT
        \STATE $z_t = f_{enc}(s_t)$ 
        \STATE $\pi(a_t|z_t ; \theta_{actor}) = f_{actor}(z_t; \theta_{actor})$ 
        \STATE Sample $a$ from $\pi(a_t|z_t ; \theta_{actor})$, get $s_{t+1}, r_t$
        \STATE $z_{t+1}\leftarrow f_{enc}(s_t+1)$
    	\STATE $t\leftarrow t + 1$, $T\leftarrow T + 1$
    \UNTIL{terminal $s_t$ \algorithmicor $t - t_{start} == t_{max}$}
    \FOR{$i\in\{t-1,\ldots,t_{start}\}$}
        \STATE $R\leftarrow r_i + \gamma R$
        \STATE $d\theta_{actor} \leftarrow d\theta_{actor} + \nabla_{\theta_{actor}} \log \pi(a_t|z_t; \theta_{actor})$
        \STATE $\qquad (R - V(z_{t}; \theta_{critic})) $
        \STATE $d\theta_{critic} \leftarrow d\theta_{critic} + \lambda_{critic}$ 
        \STATE $\qquad \nabla_{\theta_{critic}} (R - V(z_{t}; \theta_{critic}))^2$
    \ENDFOR
    \STATE Sum losses and perform asynchronous update on 
    \STATE $\qquad \theta_{actor}, \theta_{critic}$ with $d\theta_{critic}, d\theta_{actor}$
\UNTIL{$T > T_{max}$}
\end{algorithmic}
\end{algorithm}
\footnotesize{Alg.2 notes: $T_{max}$ is the total number of episodes across all threads. $\gamma$ is the discount factor for reward. $t_{max}$ is the maximum number of steps per episode.}
\end{figure}

\subsection{Rewards Learning} 
\label{sec:reward-learn}
\textit{Online, On-policy}. The rewards module is typically trained online and on-policy, using an actor-critic approach analogous to A3C~\cite{mnih2016asynchronous}, with the distinction that the actor and critic operate on the representation space $\mathcal{Z}$ built by the dynamics module. 
Algorithm \ref{alg:reward} outlines the procedure.  
In this scenario the trajectories collected by the RL agent are fed to both the dynamics and reward modules through a \textit{shared} encoder, and both modules are updated simultaneously.  In this case there is a tight dependency between the two modules: the reward module depends on the dynamics module for the representation space, whereas the the dynamics module depends on the reward module for the policy. This type of training can lead to instability, due to the non-stationary data distribution (induced by changes in the policy and changes in the encoder model that is also training). The main advantage is that as the policy improves, sample efficiency may be better and the representation space learned by the encoder can be more compact and focused on the target task.

\textit{Mixed Online/Offline}.  We consider another case, where the representation space (encoder) is static while we train the reward module.  In this case we first train the full dynamics module with sample trajectories collected offline (either off- or on-policy, as explained in Sec.~\ref{sec:dynamics-learn}) then freeze the encoder weights before training the reward module online from this fixed encoder.

\subsection{Online Planning}

A major advantage of learning the dynamics and rewards module is that at any time we can use them to perform planning in the representation space.  We feed the observation $s_t$ from our environment through our encoder to get the hidden representation $z_t$. We take an action $a_t$, and feed the action together with $z_t$ through the forward model LSTM-D to get $z_{t+1}$. We can repeat this forward sampling in the representation space to rollout full trajectories.

There are a few standard methods to choose the action $a_t$ during this procedure.  (1) We can follow a fixed given policy $\pi(z_t)$.  (2) We can exhaustively branch on the full action space, repeat to generate a tree of trajectories which terminate either at an end state or at a fixed depth. (3) We can use Monte Carlo Tree Search, which balances both exploration with efficiency to direct the tree expansion~\cite{coulom06}. In the last two cases, after expanding the tree with forward rollouts, we select the path with the maximum mean estimated value function over the trajectory to determine the next action.



\section{Modular Transfer Scenarios}
In this section, we will discuss how our architecture handles transfer to different environments and reward functions.

\subsection{Simple Generalization}
The most basic case is to train both the dynamics and reward modules from scratch and test them in the same task or environment.  In this case, we can still leverage an encoder-decoder pair learned in the same or a related task, using this as a prior on the representation space to ease sample complexity when learning the target models and policy. 

\subsection{Changes in Reward}
In this scenario the reward function changes but the state dynamics remain the same as in training. The agent now needs to learn the value function and corresponding policy according to the new reward function. Since there is no change in the state dynamics, we don't need to train the dynamics module again. We retrain the reward module in the same representation space. This is equivalent to the simple generalization case when using offline learning since the modules are decoupled.  The dynamics module is already trained off-policy, so can transfer across different reward functions if dynamics stay the same.

\subsection{Changes in Dynamics}
Now, we consider the case where the reward function and corresponding value function remain the same but the underlying dynamics change.  The state and action spaces are the same, but the mapping between them has changed.  Where we previously had an environment with dynamics described by $f_{dynamics}(s, a) = s'$, we now have a new environment described by
$g_{dynamics}(s, a) = s''$,
where $s,s',s''\in\mathcal{S}$, $a\in\mathcal{A}$ and $\mathcal{S}, \mathcal{A}$ are the same for both environments. We explore specific types of dynamics transfer in Section \ref{sec:domains}.
We need to retrain the forward model again but keep the encoder and decoder static, with the assumption that the representation learned for the prior task contains all the information necessary to transfer to the new dynamics.


\section{Experiments}
The goal of our experiments is to show that our proposed method, called DDR (Decoupled Dynamics and Reward) provides better transferability, increases ease and stability of training, and improves performance through the decoupling of dynamics and reward. We compare with the basic A3C method \cite{mnih2016asynchronous} trained from scratch for each environment as well as fine-tuned across multiple dynamics and reward functions; A3C is the most suitable baseline here due to the efficiency it achieves through parallelization, and because it can be readily applied to both continuous and discrete domains. The main trunk of our architecture is the same for our method and the baseline for fair comparison. 

We evaluate transfer ability of our module in several different scenarios: transfer a pre-trained encoder and learn the dynamics and reward, fixed reward but change in dynamics, fixed dynamics but change in reward, and change in both reward and dynamics.  We compare across fixed sample complexity $N$ with models trained from scratch.

\subsection{Continuous Control}
\label{sec:domains}
We consider the \textbf{MuJoCo} domain~\cite{TodorovET12}. Here the dynamics are defined over continuous state and action spaces.  The dynamics module must learn an intuitive physics model in order to achieve goals.  We explore multiple agents in this space---Swimmer, Hopper, Ant, and HalfCheetah---each with different dynamics.  The state spaces for these agents contain information about joint angles, joint velocities, and coordinates of the center of mass.  The reward functions are computed using the velocity of the agent and size of the action taken -- the faster the velocity and smaller the magnitude of the action taken, the larger the reward.  More detail about all the agents can be found in \cite{DBLP:journals/corr/DuanCHSA16}.

The dynamics module for experiments in this domain is trained on 100K samples generated with a random policy.  It is trained with batch size $b=512$, learning rate $\lambda=1\mathrm{e}{-3}$ for 1000 epochs with trajectories of length 20.  The reward module is trained for 1M episodes, with a maximum episode length of 500 and gradient updates every 20 steps. 
We use two linear layers apiece for the encoder and decoder, interpolated with exponential linear units (ELUs)~\cite{DBLP:journals/corr/ClevertUH15}.  Our encoder latent space is defined as $\mathcal{Z} \in \mathbb{R}^d$, with $d=200$. We also add an entropy coefficient as regularization \cite{williamsentropy} tuned for each environment, at either $1\mathrm{e}{-2}$ or $1\mathrm{e}{-3}$.
\begin{figure*}[t]
  \centering
\includegraphics[width=0.5\columnwidth]{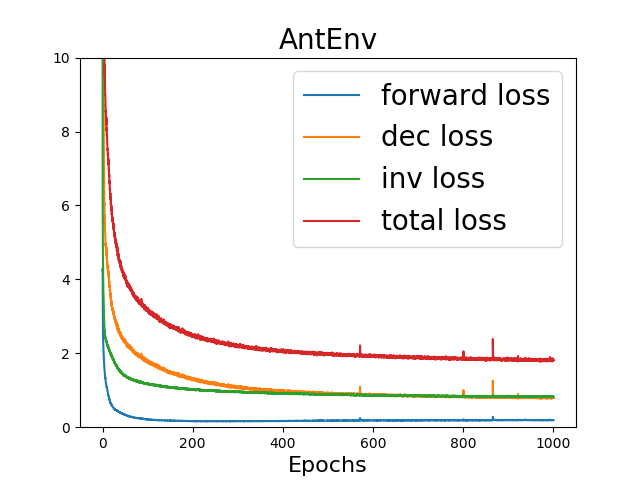}
\includegraphics[width=0.5\columnwidth]{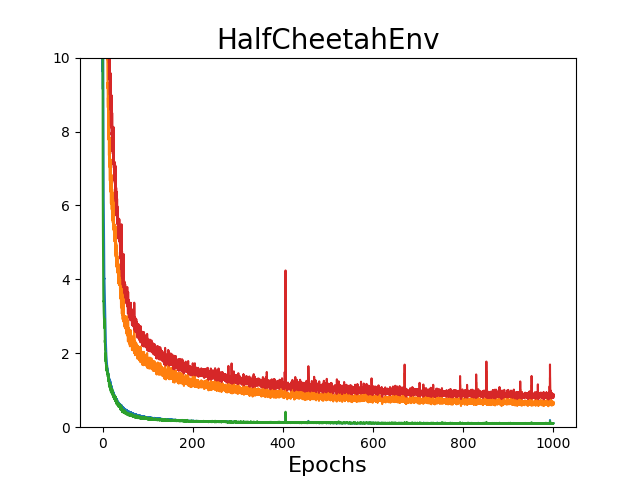}
\includegraphics[width=0.5\columnwidth]{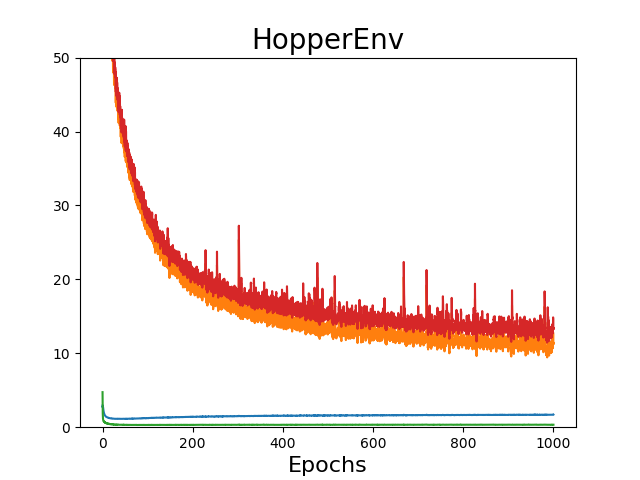}
\includegraphics[width=0.5\columnwidth]{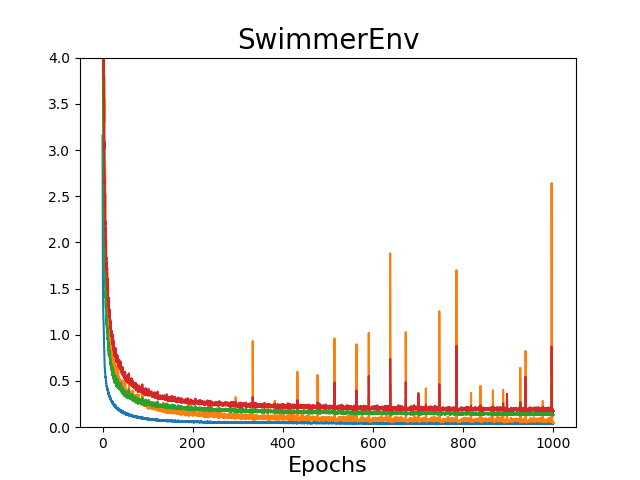}
\vskip -0.15in
  \caption{\small Dynamics losses for AntEnv, HalfCheetahEnv, HopperEnv, and SwimmerEnv.}
\label{fig:dynamics-loss}
\vskip -0.1in
\end{figure*}

\begin{figure*}[t]
  \centering
\includegraphics[width=\linewidth]{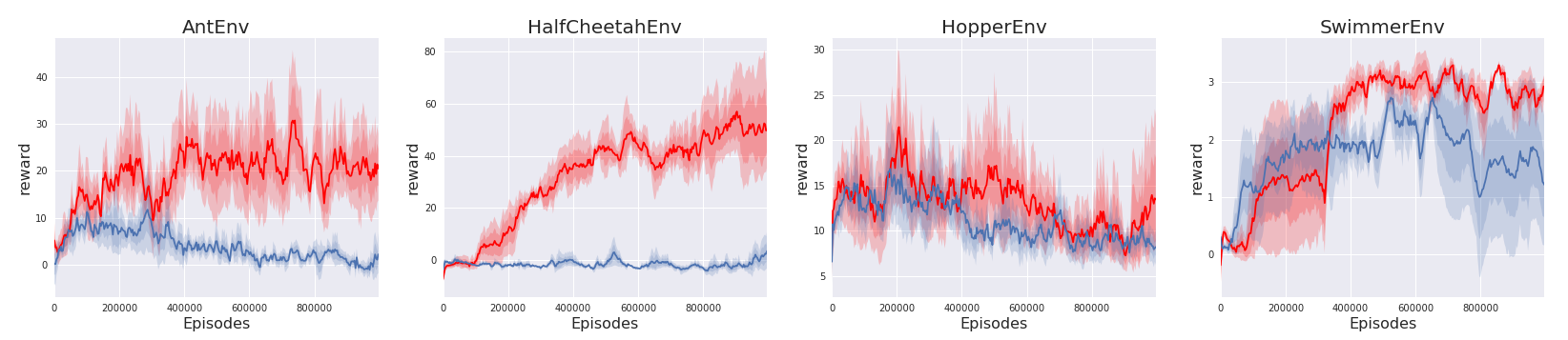}
\vskip -0.15in
  \caption{\small Rewards averaged over 5 runs for AntEnv, HalfCheetahEnv, HopperEnv, and SwimmerEnv. Red is DDR (our method), blue is A3C baseline.}
\label{fig:mujoco_rewards}
\vskip -0.15in
\end{figure*}

\begin{table}[h]
\vskip 0.1in
\begin{center}
\begin{small}
\begin{sc}
\begin{tabular}{lcccc}
\hline
\abovespace\belowspace
Task &          A3C &       DDR Prior &     DDR Online\\
\hline
\abovespace
Swimmer &       55.4 &      \textbf{68} &   19.8  \\
Ant  &          24.3 &      \textbf{508} &  54 \\
Hopper &        8 &         36 &            \textbf{38} \\
HalfCheetah &   124.8 &     \textbf{869} &  241 \\
\hline
\end{tabular}
\end{sc}
\end{small}
\end{center}
\vskip -0.1in
\caption{MuJoCo domain.  Reward averaged over 5 runs, evaluated over 100 trajectories, trained for 1M episodes.  DDR Prior is the Simple Generalization case, using a pre-trained encoder-decoder.}
\vskip -0.1in
\label{table:mujoco}
\end{table}

We first consider the Simple Generalization case. Results are presented in Table~\ref{table:mujoco}\footnote{We set a maximum episode length of 500 for evaluation. Other work does not specify the episode length used for the same environments, so our results are not directly comparable.}. For the DDR Prior case, a dynamics model is trained offline and off-policy. We then use the frozen encoder as a prior and train the reward model from scratch. For the DDR Online case, both models are trained online and on-policy as detailed in Sections \ref{sec:dynamics-learn}, \ref{sec:reward-learn}, without pre-trained components.  We observe significant performance gain from the representation transfer in 3 of the four domains; in the other case it is neutral. Our model-based approach (with or without prior) significantly outperforms standard A3C.

Next we evaluate dynamics and reward transfer. For the REWARD case, the task is modified by negating the reward given by the environment -- instead of rewarding forward velocity we reward negative velocity and train the agent to move backwards. In Table \ref{table:transfer}, for reward transfer -- the more negative the score, the better.   
We compare with a standard A3C baseline, and a variant where we fine-tune in the new environment with changed $p(\cdot|s, a)$ and/or $r(s, a)$ on top of the A3C models trained in the default environments. Note that we are only training the reward module in the new environment -- the representation space is fixed from pre-training on the original domain (with positive rewards).

For the DYNAMIC case, we increase density and damping on the joints; in this case higher reward is better. Again, the representation is pre-trained with the original domain, then we re-train the forward and inverse models in the new domain, transferring the encoder/decoder intact from the original.

Finally, we consider the case where BOTH the reward and dynamics change; once again lower reward is better. We would also like to note that the negative reward case is not symmetric to positive. The reward is computed as \textrm{reward = forward\_reward - ctrl\_cost - contact\_cost + survive\_reward}, so maximizing the negative reward is not so simple as merely maximizing negative velocity.

In all the transfer scenarios considered, the results in Table~\ref{table:transfer} show a consistent advantage for DDR which is able to leverage pre-trained modules for the components that do not change.

\begin{table}[h]
\vskip 0.1in
\begin{center}
\begin{footnotesize}
\begin{sc}
\begin{tabular}{lcccc}
\hline
\abovespace\belowspace
 Model &		    Change in & 		Change in & 	    Change in\\ 
  &		                Reward & 		Dynamics & 	        Both\\
\hline
Swimmer \\ 
\hline
\abovespace
DDR& 		    \textbf{-86.3} & 	\textbf{66.9} &     \textbf{-65}\\
A3C (f) &       0.6 & 		        50.9 &             -5.1 \\
A3C &           -4.6 &               48.8 &               -4.9 \\
\hline
Ant \\
\hline
\abovespace
DDR & 			\textbf{-908} &    \textbf{793}  &     \textbf{-366}\\ 
A3C (f)& 	    -11.8 &             50 &                -50.8 \\
A3C &           2.2 &             35.2 &               -3.5 \\
\hline
\end{tabular}
\end{sc}
\end{footnotesize}
\end{center}
\vskip -0.1in
\caption{MuJoCo Transfer Experiments. We investigate reward transfer, dynamics transfer, and both and compare with A3C finetuned with the same number of samples (A3C (f)), as well as A3C trained from scratch. Reward transfer in this case is negating the reward, so the more negative the better.}
\label{table:transfer}
\vskip -0.15in
\end{table}

\subsection{Maze Navigation}
\label{sec:maze-navigation}

Next we consider the maze navigation domain, where the task is defined over discrete state and action sets, and requires longer planning than the Mujoco domains.
In this case the agent needs to reach a goal in the least number of steps to receive maximum reward. 

\textbf{Environment.} We consider a 2D grid maze based on MazeBase~\cite{DBLP:journals/corr/SukhbaatarSSCF15}. An observation is represented as a binary vector, $s_t \in \mathbb{R}^{10 \times 10 \times 9}$, where $10 \times 10$ is the size of the grid and $9$ is the length of feature vector denoting the number of different maze elements. For the maze navigation experiments, we only have three kind of objects: Agent, Goal, and Walls. We generate the maze layout similar to the rooms domain~\cite{sutton1999between}, where the layout of the maze (position of walls) remains constant across different runs, but the agent's and the goal's location are randomly generated. The agent has $4$ primitive actions \textit{UP, DOWN, LEFT, RIGHT} that move the agent by one block in the respective direction. The agent receives a time penalty of $-0.1$ for each time-step and gets a reward of $+10.0$ on reaching the goal, after which the episode terminates. The discount factor ($\gamma$) is set to $0.99$ and the maximum episode length is set to $250$ time-steps after which the episode terminates.


\textbf{Implementation Details.} All the individual components are modeled using neural networks. The encoder ($f_{enc}$) and decoder ($f_{dec}$) are both single layer neural networks with ReLU non-linear activations~\cite{glorot2011deep}, which map the input observation to latent space $\mathcal{Z} \in \mathbb{R}^{d}$, with $d=256$. Both the LSTM-D and LSTM-R have a hidden layer with 128 units each. The Inverse model,$f_{inv}$, consists of a linear layer of size $64$ with ReLU non-linearity followed by an output layer of size $4$ with the softmax activation defining a probability over actions. The forward model, $f_{for}$, concatenates the hidden state of the LSTM-D with the one-hot encoding of the action, and passes it to a linear layer with size $256$ and tanh non-linearity. The actor and critic models, $f_{actor}$ and $f_{critic}$, both consist of a single linear layer each. 

\textbf{Pre-training.} The dynamics module is trained offline on 25K samples, each consisting of $5$ transitions, generated by following a random policy. The agent and goal location are initialized randomly at the start of each episode. The model was trained for $200$ epochs with batch size set to $100$, learning rate to $1e^{-3}$ and loss coefficients $\lambda_{dec} = 100$, $\lambda_{inv}=10$ and $\lambda_{for}=1$ respectively. The $\lambda_{for}$ was annealed linearly from $1$ to $10$ over the course of training. The reward module is trained for $10,000$ episodes, with $\lambda_{critic} = 0.5$, entropy regularization coefficient of $1e^{-3}$ and $40$ parallel asynchronous agents. 

\begin{figure}[t]
  \centering
\includegraphics[width=1.0 \columnwidth]{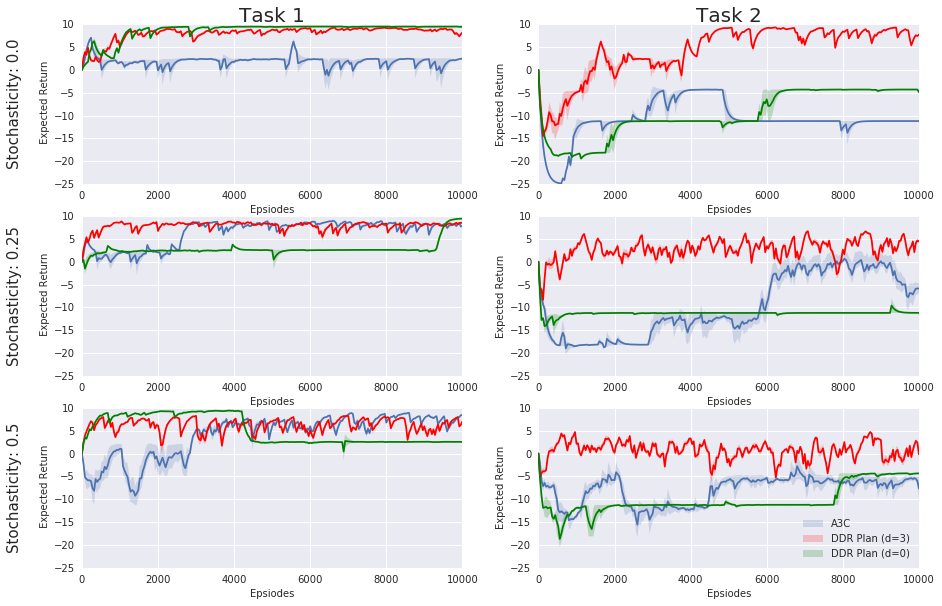}
  \caption{\small Expected reward over 5 runs on two unseen tasks with varying stochasticity in the maze domain.  Red is DDR using forward search planning with depth=$3$, green is DDR without any planning (depth=$0$), blue is A3C baseline. }
 \vskip -0.1in
\label{fig:maze_results}
\end{figure}

\textbf{Evaluation on unseen tasks.} We generate two random tasks (Task 1 and Task 2 in Fig.~\ref{fig:maze_results}), unseen by the dynamics module during training, and the goal is to learn the optimal behavior on these two tasks. We train a reward module on this new task, using the fixed pre-trained encoder to process observations and using the fixed pre-trained forward module to do online planning.  Planning is instantiated using fixed-depth forward rollouts with full branching (we could use MCTS instead).  Results in the top row of Fig~\ref{fig:maze_results} show that DDR with planning learns the new task much faster than standard A3C on both tasks. Here Task 1 is easier as the goal in the test scenario is in the same room (but different location) as in the pre-training tasks. For Task 2 we evaluate with a goal that is in a different room than the scenarios seen during pre-training of the dynamics model.

We further test the robustness of the learned dynamics module to changes in environment by introducing a stochasticity coefficient $p_s$, where $p_s$ is the probability with which the environment disregards the action taken by the agent and executes a random action. Results in Fig.~\ref{fig:maze_results} show that even if the forward model learned is imperfect (it was trained from purely deterministic actions), it still helps the agent to learn the optimal behavior faster. We also observe that it is more stable in the stochastic setting compared to A3C which exhibits significant variance. We added a model based baseline for the planning experiments in Appendix\ref{app:model_baseline}.

\subsection{Ablation of the dynamics model}
\label{sec:ablations}
 We postulate that all four components the dynamics model are necessary to learn a good representation $\mathcal{Z}$.
 To verify this, we selectively ablate components of the dynamics model. Results in Table~\ref{table:ablations} show that, as expected, we observe a significant drop in performance when removing the forward model.  Perhaps more surprisingly, we see an even greater drop in performance when removing the inverse model (but preserving the forward model). This suggests that the inverse model is essential for regularizing the dynamics problem in preventing degenerate solutions; an important finding of this work. An auto-encoder (removing both forward and inverse models) performs even worse. These results confirm that learning dynamics is crucial for a good representation space. Merely expanding out the state space creates a paradigm where it is even more difficult to learn, which is not surprising.

\begin{table}[h]
\begin{center}
\begin{small}
\begin{sc}
\begin{tabular}{lcccccc}
\hline
\abovespace\belowspace
 Agent & 		    Full & 	        No F & 	No I & 	AE \\
\hline
\abovespace
Swimmer & 	        \textbf{68} & 	25.9 &	4.48 & 	-3.3 \\
Ant  &  		    \textbf{508} &  281 & 	80.5 & 	37.5 \\
Maze Navigation &   \textbf{8.14} & 6.04 &	-0.86 & -0.85 \\
\hline
\end{tabular}
\end{sc}
\end{small}
\end{center}
\vskip -0.1in
\caption{Ablation results averaged over 5 runs. Full = All four losses, No F = No forward model, No I = No inverse model, AE = Autoencoder (no forward or inverse models).  }
\label{table:ablations}
\end{table}

\section{Related Work}
\label{related_work}
We combine ideas from many areas of previous research, especially those of successor features~\cite{dayan1993improving, DBLP:journals/corr/BarretoMSS16, Machado17b}, which also uses a decoupling mechanism in the value function to transfer across reward functions (but not dynamics)~\cite{successor_features,2016arXiv160602396K}. 

Other previous work in transfer learning includes feature sharing~\cite{NIPS2006_3143, Walsh06transferringstate} and representation learning of dynamics of the environment via predictive state representation~\cite{littman2002predictive, psr_2, psr_3}. Finally, \citet{Taylor:2009:TLR:1577069.1755839} is a comprehensive review on transfer learning in reinforcement learning that details the various types of transfer and evaluation methods.  

More specifically, we look at other model-based RL methods \cite{2016arXiv161000696F,2015arXiv150607365W} that try to transfer and plan~\cite{IJCAI07-bikram, 2015arXiv150906824X, 2016arXiv161003518C}. \citet{jaderberg2016reinforcement} proposes a state reconstruction loss and argue that having auxiliary costs help in faster learning.
\citet{2017arXiv170703497O} also performs planning strictly in the representation space.  However, they do not decouple the reward function from the dynamics function of their environment and do not learn a policy.
\citet{DBLP:journals/corr/WeberRRBGRBVHLP17} trains an environment model which learns dynamics and contains a recurrent model for imagining rollouts, but still operate in the state space.

\citet{DBLP:journals/corr/AgrawalNAML16, doi:10.1162/089976601750541778} also use an inverse model to regularize their forward model as an auxiliary task, but without decoupling, showing transfer in dynamics, and operates only in the state space.
\citet{2016arXiv161003518C} develops a transfer method specifically from simulation to real world through learning the inverse dynamics model. However, our method instead show how using both forward and inverse models as auxiliary tasks allow us to generalize and transfer with model-free policy optimization methods.

We would like to emphasize that our method is not a model-based one. We incorporate auxiliary model-based losses for learning a representation space, but do not actively use a model when training policies. This is a novel way of combining model-based and model-free methods.

\citet{DBLP:journals/corr/FinnYFAL16} is also a semi-supervised learning method that incorporates unlabeled data, but infer labels for better generalization as opposed to transfer to different reward functions.

Our model shares some similarities to the stacked LSTM used in \citet{2016arXiv161105763W}, but our decoupling strategy enables more efficient training and more modular transfer.

\section{Discussion}
\label{conclusion}

We present a decoupled model-based RL framework that offers efficient and modular reuse to pre-trained models and enables robust transfer across tasks.  There are several key ingredients to this approach. By learning an encoder jointly with the dynamics we can focus representation on relevant information. The pre-training of a forward model, enables planning which leads to faster policy optimizing.  The incorporation of an inverse model has an important stabilizing effect on the dynamics model.  The modularity of the rewards model allows off- and on-policy learning. Finally, the approach can be used for both discrete and continuous domains.

Throughout our experiments we consistently observe that the offline, decoupled mode of training significantly outperforms that of online/on-policy training.  One of the advantages of training modules in a decoupled manner is that the dynamics learning becomes a supervised learning task, and converges faster and more stably than when both modules are trained simultaneously.  This leaves open the question of how to effectively train dynamics models in an on-policy setting that isn't as volatile as the online version explored here.  One possibility is to incorporate additional supervision (e.g. adding intrinsic motivation that is relevant to a specific family of tasks), another possibility is to explore mechanisms to directly stabilize the non-stationarity of the data distribution (e.g. as when using target networks in DQN \cite{mnih2015human}).

Our results also highlight the brittleness of standard A3C, which despite its popularity (due to fast training enabled by parallelization) performs poorly on several tasks. We provide results from our hyperparameter search on A3C in Appendix \ref{app:hyperparam_a3c}. In many ways, our core contributions are orthogonal to the policy optimization method, and DDR could be extended to other policy optimization methods such as TRPO \cite{DBLP:journals/corr/SchulmanLMJA15} or PPO \cite{DBLP:journals/corr/SchulmanWDRK17}, results for which are in Appendix \ref{app:ppo_results}, where we would expect to still see large gains in performance and transferability due to the modularity of the architecture.


\section*{Acknowledgements}
Thanks to Alessandro Lazaric and Ryan Lowe for helpful feedback.

\bibliography{paper}
\bibliographystyle{icml2017}

\clearpage

\appendix

\section{Hyperparameter Search for A3C}
\label{app:hyperparam_a3c}
We did a hyperparameter search over learning rate, value coefficient, entropy regularization, and with and without generalized advantage estimation (GAE) for all MuJoCo environments with vanilla A3C and chose the hyperparameters that performed best. We then carried over those hyperparameters to DDR for the reward module and performed some tuning on the loss weights for the various auxiliary losses -- forward loss, inverse loss, and decoder loss in the dynamics module.

\section{PPO Results}
\label{app:ppo_results}
\begin{table}[h]
\vskip 0.1in
\begin{center}
\begin{small}
\begin{sc}
\begin{tabular}{lcccc}
\hline
\abovespace\belowspace
Task &          PPO &       DDR Prior \\
\hline
\abovespace
Swimmer &       49.22 &      \textbf{76.22} \\
Ant  &          85.12 &      \textbf{312}\\
Hopper &        299.6 &         \textbf{337.2} \\
HalfCheetah &  482.6  &     \textbf{1902.8} \\
\hline
\end{tabular}
\end{sc}
\end{small}
\end{center}
\vskip -0.1in
\caption{MuJoCo domain.  Reward averaged over 5 runs, evaluated over 100 trajectories, trained for 1M episodes.  DDR Prior is the Simple Generalization case, using a pre-trained encoder-decoder.}
\vskip -0.1in
\label{table:mujoco-ppo}
\end{table}

\newpage
\section{Comparison with Model Based Baseline}
\label{app:model_baseline}

\begin{figure}[h]
  \centering
\includegraphics[width=1.0\columnwidth]{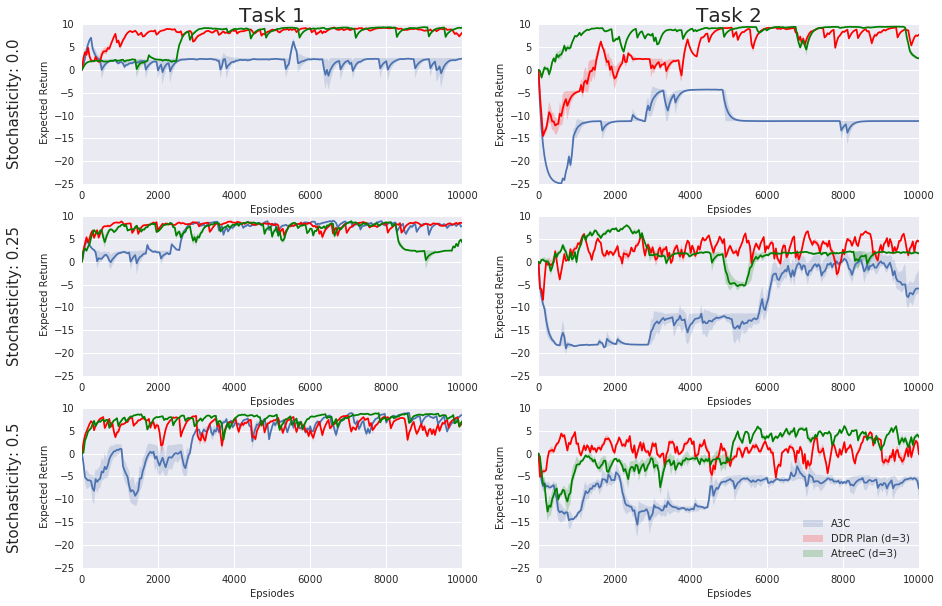}
  \caption{\small Expected reward over 5 runs on two unseen tasks with varying stochasticity in the 4-rooms maze domain.  Both DDR and ATreeC use exhaustive forward search planning with depth=$3$, blue is A3C baseline. }
 \vskip -0.1in
\label{fig:model_baseline}
\end{figure}

We compare our approach with a model based baseline for the planning experiments, an asynchronous version of ATreeC \cite{farquhar2017treeqn}, that combines model-free reinforcement learning with on-line planning. Results in Fig.\ref{fig:model_baseline} show that both our method (DDR planning) and ATreeC achieve comparable performance and do better than A3C, especially in the stochastic environments. We use the same setting as in Section~\ref{sec:maze-navigation}. We have tried to keep the comparison fair with equal capacity and designs, extensive hyper-parameter search, averaged over 5 runs and using the same depth for exhaustive search for the planning. However, ATreeC is not designed for representation transfer, which is one of the main advantages of our proposed DDR (highlighted in other experiments).

\end{document}